\documentclass[twocolumn]{article}
\usepackage{authblk}
\usepackage{times}
\usepackage{soul}
\usepackage{url}
\usepackage[hidelinks]{hyperref}
\usepackage[utf8]{inputenc}
\usepackage[small]{caption}
\usepackage{graphicx}
\usepackage{amsmath}
\usepackage{amsthm}
\usepackage{booktabs}
\usepackage{algorithm}
\usepackage{algorithmic}
\usepackage[switch]{lineno}

\usepackage{array}
\usepackage{enumitem}
\usepackage{float}
\usepackage{listings}
\usepackage{multicol}
\usepackage{multirow}
\usepackage{natbib}
\usepackage{tikz}
\usetikzlibrary{patterns}
\usetikzlibrary{shapes.geometric}

\title{Logic.py: Bridging the Gap between LLMs and Constraint Solvers}

\author[1]{Pascal Kesseli}
\author[1]{Peter O'Hearn}
\author[1]{Ricardo Silveira Cabral}
\affil[1]{Meta}
\affil[ ]{\textit {\{pkesseli, peteroh, rdsc\}@meta.com}}

\newcommand{\logiclang}{\textit{Logic.py}}

\definecolor{spruce}{rgb}{0.110, 0.390, 0.255}
\definecolor{lightgreen}{rgb}{0.56, 0.93, 0.56}
\definecolor{lightblue}{rgb}{0.68, 0.85, 0.90}

\lstdefinestyle{c}{
  language=C,
  basicstyle=\ttfamily,
  keywordstyle=\color{blue},
  commentstyle=\color{spruce},
  stringstyle=\color{red},
  breaklines=true,
  tabsize=2
}

\lstdefinestyle{python}{
  language=Python,
  basicstyle=\ttfamily,
  keywordstyle=\color{blue},
  commentstyle=\color{spruce},
  stringstyle=\color{red},
  breaklines=true,
  tabsize=2
}

\lstdefinelanguage{smt}{
  keywords={forall, implies, distinct, not, =}
}

\begin{document}

\date{2025-02-17}
\maketitle

\begin{abstract}
    We present a novel approach to formalise and solve search-based problems
    using large language models, which significantly improves upon previous
    state-of-the-art results. We demonstrate the efficacy of this approach on
    the logic puzzles benchmark ZebraLogicBench. Instead of letting the LLM
    attempt to directly solve the puzzles, our method prompts the model to
    formalise the problem in a logic-focused domain-specific language (DSL)
    called \logiclang. This formalised representation is then solved using a
    constraint solver, leveraging the strengths of both the language model and
    the solver. Our approach achieves a remarkable 65\% absolute improvement
    over the baseline performance of Llama 3.1 70B on ZebraLogicBench, setting a
    new state-of-the-art with an accuracy of over 90\%. This significant
    advancement demonstrates the potential of combining language models with
    domain-specific languages and auxiliary tools on traditionally challenging
    tasks for LLMs.
\end{abstract}

\section{Introduction}

Large language models have revolutionised the field of natural language
processing, achieving state-of-the-art results in various tasks such as language
translation, text summarisation, and question answering. However, despite their
impressive performance, LLMs have historically struggled with certain tasks that
require a deeper understanding of mathematical and logical concepts. For
instance, \cite{kambhampati2024llmscantplanhelp} demonstrated that LLMs are
unable to plan and reason about complex problems, highlighting the need for
further research in this area. In this paper, we focus on improving the
performance of LLMs in solving Logic Grid Puzzles, also called Zebra Logic
Puzzles or Einstein's Riddles, which we explain in more detail in
Sec.~\ref{logic-grid-puzzles}. We present the following research contributions:

\begin{enumerate}
    \item \textbf{\logiclang}: We introduce a domain-specific language called
    \logiclang\, which facilitates expressing logic and search-based problems by
    LLMs.
    \item \textbf{Logic Agent}: We implement an agentic solver engine which
    accepts search-based, informal problem statements, formalises them in
    \logiclang\ and solves them using a constraint sovler.
    \item \textbf{ZebraLogicBench Evaluation}: We evaluate the efficacy of this
    approach on the logic puzzle benchmark
    \textit{ZebraLogicBench}~\cite{zebralogic2024}.
\end{enumerate}

\subsection{Related Work}

Prior research has explored techniques to enhance the ability of LLMs in these
central reasoning tasks, such as chain-of-thought prompting and introducing
symbolic representations. However, according to
\cite{berman2024solvingzebrapuzzlesusing}, these frameworks often struggle with
complex logical problems like Zebra puzzles, partly due to the inherent
difficulty of translating natural language clues into logical statements. They
propose integrating LLMs with theorem provers to tackle such challenges,
demonstrating significant improvements in puzzle-solving capabilities.

\cite{al-negheimish2023augmenting} highlight that the challenge of numerical
reasoning in machine reading comprehension has been addressed by various
prompting strategies for LLMs. However, these approaches often struggle to
provide robust and interpretable reasoning. They contrast these techniques
against their neuro-symbolic approach, which has shown promising results by
decomposing complex questions into simpler ones and using symbolic learning
methods to learn rules for recomposing partial answers. This approach boasts
data efficiency and facilitates robust numerical reasoning with interpretable
and verifiable reasoning traces.

The LLM-Augmenter system, proposed by \cite{peng2023checkfactstryagain},
addresses the limitations of LLMs in real-world applications by augmenting them
with plug-and-play modules that ground responses in external knowledge and
iteratively revise prompts to improve factuality. Similarly, the Logic-Enhanced
Language Model Agents (LELMA) framework, proposed by
\cite{mensfelt2024logicenhancedlanguagemodelagents}, integrates LLMs with
symbolic AI to enhance the trustworthiness of social simulations, addressing
issues such as hallucinations and logical inconsistencies through logical
verification and self-refinement.

\cite{imani2023mathpromptermathematicalreasoningusing} propose a technique to
improve the performance of LLMs on arithmetic problems by generating multiple
algebraic expressions or Python functions to solve the same math problem in
different ways, thereby increasing confidence in the output results. The
Toolformer self-supervised model, presented in
\cite{schick2023toolformerlanguagemodelsteach}, enables language models to
leverage external tools via simple APIs, thereby enhancing their performance on
various downstream tasks. \cite{fedoseev2024llm} propose a method to enhance
LLMs mathematical problem-solving capabilities by fine-tuning them on a dataset
of synthetic problems and solutions generated using Satisfiability Modulo
Theories (SMT) solvers, specifically the Z3 API.

Logic-LM is a framework that combines LLMs with symbolic solvers to enhance
logical reasoning capabilities, demonstrating substantial performance
improvements over traditional LLM-based approaches \cite{pan-etal-2023-logic}.
\cite{ye2023satlmsatisfiabilityaidedlanguagemodels} introduce
Satisfiability-aided Language Modeling (SatLM), a method that combines large
language models with automated theorem provers to enhance reasoning
capabilities, demonstrating state-of-the-art performance on multiple datasets.

From the work cited above, \cite{pan-etal-2023-logic},
\cite{ye2023satlmsatisfiabilityaidedlanguagemodels}, and
\cite{berman2024solvingzebrapuzzlesusing} present techniques addressing problem
domains most comparable to the Logic Grid Puzzles which we explore in this
paper. Similar to the approach we present, formalisation and the use of
auxiliary tools are a shared feature of this category of research. An important
distinguishing aspect of our work is the use of a logic-focused domain-specific
language, which we present in more detail in Sec.~\ref{logiclang}.

The general case for a DSL is not unlike that for intermediate languages in compilation. Given a new programming language it is ``obvious'' from the Church-Turing thesis that a mapping to assembly language exists, but is not necessarily obvious if a mapping with good efficiency properties (say) exists, and the compilation research community has found it helpful to use intermediate languages to structure the design of compilers. Here, it might seem likely that constraint solvers (automatic theorem provers) could be helpful to approach logic problems stated in natural language, but just how helpful or the best way to do so is not a priori obvious; like in compilation, our thesis is that DSL's can help. A similar analogy is the relation between SQL and first-order logic; although SQL provides facilities than make for briefer or more direct human expression than their expansions into FOL. A similar example for Logic.py is presented in Section \ref{type-decorators}.

Our results
support the case that this DSL can be helpful in structuring the mapping from natural language to that of a solver. In particular, \cite{berman2024solvingzebrapuzzlesusing} evaluate their
approach against a benchmark set of 114 Zebra puzzles. Their multi-agent system has a more sophisticated translation process which includes a refinement loop, but they
raise the puzzle accuracy of GPT-4 from 23.7\% to 55.3\%, compared to our
improvement from 24.9\% to 91.4\%.

\subsection{Logic Grid Puzzles}
\label{logic-grid-puzzles}

We evaluate the effectiveness of our approach on the \textit{ZebraLogicBench}
benchmark presented in \cite{zebralogic2024}. \textit{ZebraLogicBench} is a
dataset of 1000 Logic Grid Puzzles, also referred to as Zebra Puzzles. These
puzzles consist of a series of clues about features of entities in a described
environment. In order to solve the puzzle, one has to guess the correct features
of all entities, while respecting all the information provided in the cluses.
Fig.~\ref{zebra-puzzle-description} gives an example of sucha a description of
entities and their features in a Zebra puzzle.

\begin{figure}[htbp]
    \centering
    \begin{minipage}{\linewidth}
        There are 4 houses, numbered 1 to 4 from left to right, as seen from across the street. Each house is occupied by a different person. Each house has a unique attribute for each of the following characteristics:
        \begin{itemize}
            \item Each person has a unique name: Alice, Eric, Arnold, Peter
            \item Each person has an occupation: artist, engineer, teacher, doctor
            \item People have unique favorite book genres: fantasy, science fiction, mystery, romance
            \item People use unique phone models: google pixel 6, iphone 13, oneplus 9, samsung galaxy s21
        \end{itemize}
    \end{minipage}
    \caption{Description of the Zebra logic puzzle}
    \label{zebra-puzzle-description}
\end{figure}

Consequently, Fig.~\ref{zebra-puzzle-clues} provides a set of example clues
expressed over these entities.

\begin{figure}[htbp]
    \centering
    \begin{lstlisting}[language=, breaklines=true]
  Clues:
  1. The person who is an engineer is directly left of the person who uses a Samsung Galaxy S21.
  2. The person who loves fantasy books is in the second house.
  3. Alice is not in the second house.
  4. Eric is the person who is a
     teacher.
  5. The person who uses a Samsung Galaxy S21 is the person who loves fantasy books.
  6. The person who uses an iPhone 13 is the person who loves science fiction books.
  7. The person who loves science fiction books is somewhere to the left of the person who uses a OnePlus 9.
  8. The person who uses a OnePlus 9 is Arnold.
  9. The person who is a doctor is the person who loves mystery books.
  10. The person who uses an iPhone 13 is the person who is a teacher.
    \end{lstlisting}
    \caption{Clues for the Zebra logic puzzle}
    \label{zebra-puzzle-clues}
\end{figure}

\textit{ZebraLogicBench} categorises puzzles into separate difficulties
according to the number of entities and distinct features the LLM needs to
reason about in order to solve it.

Puzzles conforming to the following shapes are considered easy puzzles:
\begin{multicols}{4}
    \begin{itemize}
        \item $2 \times 3$
        \item $2 \times 4$
        \item $2 \times 5$
        \item $2 \times 6$
        \item $3 \times 2$
        \item $3 \times 3$
    \end{itemize}
\end{multicols}

Puzzles of the following shapes are categorised as hard puzzles:
\begin{multicols}{4}
    \begin{itemize}
        \item $3 \times 4$
        \item $3 \times 5$
        \item $3 \times 6$
        \item $4 \times 2$
        \item $4 \times 3$
        \item $4 \times 4$
        \item $4 \times 5$
        \item $4 \times 6$
        \item $5 \times 2$
        \item $5 \times 3$
        \item $5 \times 4$
        \item $5 \times 5$
        \item $5 \times 6$
        \item $6 \times 2$
        \item $6 \times 3$
        \item $6 \times 4$
        \item $6 \times 5$
        \item $6 \times 6$
    \end{itemize}
\end{multicols}

Consequently, the $4 \times 4$ example puzzle illustrated in
Fig.~\ref{zebra-puzzle-description} and Fig.~\ref{zebra-puzzle-clues} would be
classified as a hard puzzle, but at the lower end of the difficulty specturm. A
valid solution of this particular puzzle is illustrated in
Tab.~\ref{zebra-puzzle-solution}.

\begin{table}
    \resizebox{\linewidth}{!}{
        \begin{tabular}{|c|c|c|c|c|}
            \hline
            House & Name & Occupation & BookGenre & PhoneModel \\
            \hline
            1 & Alice & Engineer & Romance & Pixel 6 \\
            2 & Peter & Artist & Fantasy & Galaxy S21 \\
            3 & Eric & Teacher & SciFi & iPhone 13 \\
            4 & Arnold & Doctor & Mystery & OnePlus 9 \\
            \hline
        \end{tabular}
    }
    \caption{Zebra Puzzle Example Solution}
    \label{zebra-puzzle-solution}
\end{table}

\subsection{CBMC - Bounded Model Checker for C and C++ programs}

CBMC~\cite{ckl2004} is a static analysis tool designed for C and C++ programs.
It operates by mapping programs to formulas in a back-end solver, typically SAT
or SMT, which are satisfiable if and only if the mapped program exhibits a
specific property. This capability enables CBMC to effectively check programs
for bugs or other properties of interest. Notably, CBMC is powered by the
underlying CPROVER static analysis engine, which also supports other language
front-ends, such as the JBMC~\cite{ckkst2018} Java front-end.

Since CBMC implements a mapping between programs and SAT or SMT formulas, it can
serve as a convenient front-end for such constraint sovlers, exposing an API
that allows expressing SAT or SMT formulas as C programs with free input
variables. Expressing formulas in this fashion makes many constraint solver
tasks more accessible for human developers, and it is a core hypothesis of this
paper that it equally simplifies the use of constraint solvers for LLMs. We
describe this DSL that we expose to the LLM in more detail in
Sec.~\ref{logiclang}.

\section{The \logiclang\ language}
\label{logiclang}

Our goal in designing \logiclang\ was to provide a streamlined API language that
allows an LLM to efficiently express search-based problems and leverage a
constraint solver. We optimised for the folowing criteria:
\begin{enumerate}
    \item \textbf{Robustness}: The language should minimise the surface area for
    syntax errors.
    \item \textbf{Conciseness}: The model should be able to express the
    necessary constraints to solve a search-based problem without boilerplate or
    needing to worry about unrelated implementation details of the programming
    language.
    \item \textbf{Expressiveness}: While common constraints, such as uniqueness
    of a property, should be easy to express in our DSL, the language must not
    be restricted to just these common cases. Instead, it must allow the LLM to
    express arbitrary constraints in the underlying constraint solver framework
    if necessary.
\end{enumerate}

To provide a good basis for our DSL in terms of robustness and conciseness, we
decided to not start with CBMC's native C as a base language for \logiclang\,
but instead, as its name suggests, we selected Python for this purpose. We use
libCST\footnote{https://github.com/Instagram/LibCST} to transform \logiclang\
to C for analysis by CBMC, as explained in more detail in
Sec.~\ref{architecture}. Python allows the model to introduce new variables
without the need for explicit, typed declarations. Similarly, these variables
can be reused for other values later on without needing to worry about type
compatiblity.

\subsection{Type Decorators}
\label{type-decorators}

In order to further improve the conciseness of \logiclang, we borrow well-known
language features from existing languages and combine them in our DSL. As an
example, expressing uniqueness of a property in SMT directy requries code
similar to the example in Fig.~\ref{uniqueness-smt}.

\begin{figure}[htbp]
    \centering
    \begin{lstlisting}[language=smt]
(forall ((x T) (y T))
    (=> (distinct x y)
        (not
            (= (id x) (id y))))
    \end{lstlisting}
    \caption{Uniqueness constraint in SMT}
    \label{uniqueness-smt}
\end{figure}

The snippet in Fig.~\ref{uniqueness-smt} requires that two distinct objects
should not have the same \textit{id} property. Compare this to expressing the
same constraint in the Data Definition Language (DDL) of the Structured Query
Language (SQL) in Fig.~\ref{uniqueness-sql}.

\begin{figure}[htbp]
    \centering
    \begin{lstlisting}[language=sql]
id INT UNIQUE
    \end{lstlisting}
    \caption{Uniqueness constraint in SQL}
    \label{uniqueness-sql}
\end{figure}

Motivated by this example, we introduced a set of custom type decorators to
\logiclang\, which allow to express common property constraints in a much
shorter, less error-prone way. We provide full list of \logiclang\ type decorators
in Tab.~\ref{logiclang-type-decorators}.

\begin{table}
    \renewcommand{\arraystretch}{1.5}
    \resizebox{\linewidth}{!}{
        \begin{tabular}{|l|p{16em}|}
            \hline
            Decorator & Description \\
            \hline
            $Unique[T]$ & No two objects of the same type can have the same value
            for a property marked $Unique$. The behaviour is equivalent to the SQL
            \textit{UNIQUE} keyword. \\
            \hline
            $Domain[T, D]$ & Restricts the possible values of the property of type
            $T$ to the values in $D$, where $D$ is either a sequence of values, or a
            numerical range. This a more concise way of expressing value domains in
            the CPROVER framework using assumptions. \\
            \hline
            $list[T, S]$ & Creates a fixed-size list of type `T' and size `S'. This
            is equivalent to fixed size arrays in ANSI C. \\
            \hline
        \end{tabular}
    }
    \caption{\logiclang\ type decorators}
    \label{logiclang-type-decorators}
\end{table}

\subsection{Free Variables, Assumptions, and Assertions}

A key concept in using constraint solvers are free variables. By default, any
variable which is not explicitly initialised in \logiclang\ is assumed to be a
free variable. Consequently, the constraint solver is allowed to assign it any
value in order to find a satisfying assignment for all the specified
constraints. We also say that such a variable has a nondeterministc value.

The CPROVER framework also allows users to specify assumptions and assertions.
These two features work in tandem: Assumptions act as preconditions, usually
expressed over free variables, to constrain the solver's search to valid or
interesting inputs. Assertions on the other hand represent safety properties,
for which the solver tries to find a falsifying assignment in order to prove the
presence of a bug. Both of these features are exposed in \logiclang\ via the
$assume(...)$ function and the $assert ...$ statement, respectively.

\begin{table}
    \renewcommand{\arraystretch}{1.5}
    \resizebox{\linewidth}{!}{
        \begin{tabular}{|l|p{13em}|}
            \hline
            Feature & Description \\
            \hline
            Uninitialised variables & Uninitialised variables receive a
            nondeterministic value. When solving search-based problems, the solution
            is automaically marked nondeterministic by our engine and the model can
            constraint the solution according to its requirements.  \\
            \hline
            $assume(pred)$ & Constrains search paths for solutions to only the
            values where $pred$ is true. \\
            \hline
            $assert\ pred$ & In static analysis verification scenarios, assertions
            must be true for all possible inputs satisfying assumptions. In our
            search-based problem harness, they are equivalent to assumptions.\\
            \hline
            $nondet(list)$ & Returns a nondeterministic but valid element from
            $list$. This can be used in combination with assumptions to find
            elements in a list that satisfy a predicate, then express additional
            constraints about them. \\
            \hline
        \end{tabular}
    }
  \caption{\logiclang\ nondet features}
  \label{logiclang-nondet-features}
\end{table}

It should be noted that in \logiclang, from the perspective of the LLM, these
two become interchangeable. We interpret assertions specified by the model not
as safety properties to be checked for violations, but instead as requirements
on a valid solution. Interpreting assertions as assumptions and passing only a
single reachability assertion to the constraint solver is a common pattern in
program synthesis use cases~\cite{program-synthesis-for-program-analysis}. We
will explain the structure of constraints we produce at CPROVER intermediate
representation level in more detail in Sec.~\ref{architecture}. We summarise the
nondeterminism-related \logiclang\ features in
Tab.~\ref{logiclang-nondet-features}.

\section{Search Problem Formalisation}
\label{search-problem-formalisation}

After reviewing the features of \logiclang\ in Sec.~\ref{logiclang}, we now
review how we prompt the model to express search problems in this language. We
provide the full prompts in our open source agent project
\emph{polymath}\footnote{https://github.com/facebookresearch/polymath}. Note
that all our prompts are zero-shot in that we do not provide any full Zebra
puzzle as example to the model. Instead, we only explain \logiclang\ and how to
use its nondeterministic features to express generic search-based problems.

\subsection{Describing a Result Data Structure}

Before attempting to convert the required properties and constraints for the
solution of a search-based problem, we first prompt the model to define the data
structure that can contain such a solution. We ask the model to define this data
structure in \logiclang\ and be as precise as possible with respect to type
annotations. This constrains the space of valid solutions purely based on the
domain of the problem, irrespective of explicit constraints for now.
Fig.~\ref{data-structure-example} provides an example of the kind of data
structure the LLM will generate.

\begin{figure}
    \begin{lstlisting}[style=python]
class House:
  house_number: Unique[
    Domain[int, range(1, 7)]
  ]
  name: Unique[
    Domain[str, "Alice", "Eric", ...]
  ]
  # ...

class PuzzleSolution:
  houses: list[House, 6]

    \end{lstlisting}
    \caption{Model Output Example: Result Data Structure}
    \label{data-structure-example}
\end{figure}

\subsection{Constraining a Correct Solution}

Once the data structure is defined, we prompt the model to generate a validation
function that accepts an argument of this type and asserts that it is the
correct solution we are searching for. In the case of Zebra puzzles, this leads
to the model adding assertions corresponding to the clues in the puzzle. This
approach transforms the challenge for the LLM fundamentally: Instead of
searching the solution space for configurations that satisfy the clues stated in
the puzzle, it just needs to be able to reason about the clues themselves.
Fig.~\ref{constraints-example} shows a few example clues and how they can be
formalised in a \logiclang\ data strucutre chosen by the model.

\begin{figure}
    \begin{lstlisting}[style=python]
def validate(solution: PuzzleSolution) -> None:
  # Clue 1: Bob is the person who uses
  # a Xiaomi Mi 11.
  bob = nondet(solution.houses)
  assume(bob.name == "Bob")
  assert bob.phone == "xiaomi mi 11"

  # Clue 2: The person who loves the
  # soup is in the fourth house.
  soup_lover = nondet(solution.houses)
  assume(soup_lover.lunch == "soup")
  assert soup_lover.house_number == 4

  # Clue 3: The Dragonfruit smoothie
  # lover is somewhere to the left of
  # the person in a ranch-style home.
  d = nondet(solution.houses)
  assume(d.smoothie == "dragonfruit")
  r = nondet(solution.houses)
  assume(r.house_style == "ranch")
  assert d.house_number < r.house_number

  # ...

    \end{lstlisting}
    \caption{Model Output Example: Solution Constraints}
    \label{constraints-example}
\end{figure}

\section{Logic Agent Architecture}
\label{architecture}

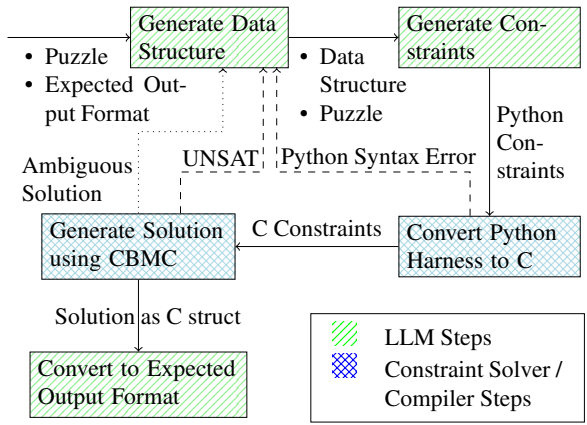
\begin{figure}
    \resizebox{\linewidth}{!}{
        \begin{tikzpicture}
            % Labelled nodes
            \node (data-structure) [draw, rectangle, text width=6em, pattern=north east lines, pattern color=lightgreen] {Generate Data Structure};
            \node (constraints) [draw, rectangle, text width=7em, right of=data-structure, xshift=9em, pattern=north east lines, pattern color=lightgreen] {Generate Constraints};
            \node (python-to-c) [draw, rectangle, text width=7em, below of=constraints, yshift=-6em, pattern=crosshatch, pattern color=lightblue] {Convert Python Harness to C};
            \node (cbmc) [draw, rectangle, text width=7.5em, left of=python-to-c, xshift=-12em, pattern=crosshatch, pattern color=lightblue] {Generate Solution using CBMC};
            \node (format) [draw, rectangle, text width=8.5em, below of=cbmc, yshift=-3em, pattern=north east lines, pattern color=lightgreen] {Convert to Expected Output Format};

            % Error path intermediate coordinates
            \coordinate (python-syntax-error) at (1, -2) {};
            \coordinate (unsat) at (0.8, -2) {};
            \coordinate (ambiguous) at (0.2, -1.4) {};

            % Legend
            \node[anchor=north west, draw, fill=white] (legend) at (1.5, -4.1) {
                \begin{tabular}{lp{7.5em}}
                    \tikz{\fill[pattern=north east lines, pattern color=green] (0, 0) rectangle (1em, 1em);} & LLM Steps \\
                    \tikz{\fill[pattern=crosshatch, pattern color=blue] (0, 0) rectangle (1em, 1em);} & Constraint Solver / Compiler Steps
                \end{tabular}
            };

            % Edges
            \draw[->] (-3, 0) -- node[xshift=1em, yshift=-2.1em] {
                \begin{minipage}{8em}
                    \begin{itemize}[noitemsep]
                        \item Puzzle
                        \item Expected Output Format
                    \end{itemize}
                \end{minipage}
            } (data-structure.west);
            \draw[->] (data-structure) -- node[yshift=-2em] {
                \begin{minipage}{6em}
                    \begin{itemize}[noitemsep]
                        \item Data Structure
                        \item Puzzle
                    \end{itemize}
                \end{minipage}
            } (constraints);
            \draw[->] (constraints) -- node[text width=1em, xshift=0.8em] {Python Constraints} (python-to-c);
            \draw[->] (python-to-c) -- node[yshift=0.8em] {C Constraints} (cbmc);
            \draw[->] (cbmc) -- node[xshift=0.4em] {Solution as C struct} (format);

            \draw[->, dashed] ([xshift=-2em]python-to-c) |- (python-syntax-error) |- node[xshift=4.3em, yshift=-3.8em] {Python Syntax Error} (python-syntax-error |- data-structure.south);
            \draw[->, dashed] ([xshift=4em]cbmc) |- (unsat) |- node[xshift=-1.8em, yshift=-3.8em] {UNSAT} (unsat |- data-structure.south);
            \draw[->, dotted] (cbmc) |- (ambiguous) |- node[text width=4em, xshift=-6.5em, yshift=-4.7em] {Ambiguous Solution} (ambiguous |- data-structure.south);
        \end{tikzpicture}
    }
    \caption{Logic Agent Architecture}
    \label{logic-agent-architecture}
\end{figure}

Fig.~\ref{logic-agent-architecture} illustrates the full implementation
architecture of our solver engine, starting with the formalisation steps
outlined in Sec.~\ref{search-problem-formalisation}.

Our engine converts these \logiclang\ constraints into an equivalent,
lower-level C representation using a libCST transformer. During this process,
the type decorators introduced in Sec.~\ref{type-decorators} are mapped to
matching initialisation helpers in CBMC's IR. An example of this mapping is
illustrated in Fig.~\ref{cbmc-data-structure-example}.

\begin{figure}
    \begin{lstlisting}[style=c]
struct House {
  int house_number;
  const char * name;
  const char * smoothie;
  // ...
};

static int House_house_number[] =
  {1, 2, 3, 4, 5, 6};
static bool House_house_number_used[6];
static const char * House_name[] =
  {"Alice", "Eric", "Peter", ...};
static bool House_name_used[6];
// ...

#define __CPROVER_unique_domain( \
  field, field_domain_array) \
{ \
  size_t index; \
  __CPROVER_assume(index < \
    (sizeof(field_domain_array) / \
     sizeof(field_domain_array[0]))); \
  __CPROVER_assume( \
    !field_domain_array##_used[index]); \
  field_domain_array##_used[index] = \
    true; \
  field = field_domain_array[index]; \
}
// ...

static void init_House(
  struct House * instance) {

  __CPROVER_unique_domain(
    instance->house_number,
    House_house_number
  );
  __CPROVER_unique_domain(
    instance->name,
    House_name
  );
  // ...
}

    \end{lstlisting}
    \caption{CBMC Data Structure Example}
    \label{cbmc-data-structure-example}
\end{figure}

The validation function containing the constraints derived from the Zebra Logic
clues is equally converted to a C representation, and embedded into a search
harness. In this harness, a nondeterministic instance of the result data
structure proposed by the model is initialised using the type decorator
information, then constrained using the converted validation function. All
assertions in the validation function are converted into assumptions, and we add
a single reachabilty assertion to prompt the constraint solver to find an
assignment for the nondeterministc puzzle solution such that it satisifes all
constraints. An example harness is shown in
Fig.~\ref{cbmc-search-constraint-example}.

\begin{figure}
    \begin{lstlisting}[style=c]
static void validate(
  struct PuzzleSolution solution) {

  bob = __CPROVER_nondet_element(
    solution.houses);
  __CPROVER_assume(bob.name == "Bob");
  __CPROVER_assume(bob.phone ==
    "xiaomi mi 11");
  // ...
}

// ...

int main(void) {
  struct PuzzleSolution solution;
  init_PuzzleSolution(&solution);
  validate(solution);

  __CPROVER_output("solution",
    solution);
  __CPROVER_assert(false, "");
}

    \end{lstlisting}
    \caption{CBMC Search Constraint Example}
    \label{cbmc-search-constraint-example}
\end{figure}

\subsection{Error Recovery}

Due to the heuristic nature of the formalisation and constraint solver steps,
our solver engine can fail at various steps along the pipeline. The model might
produce invalid \logiclang\ code to begin with, leading to syntax errors in
libCST. These errors are caught during the C harness generation, in which case
we revert to the original data structure generation step and start the process
from scratch. We currently do not provide any information about the syntax error
to the model or ask it to fix its prior mistakes, and such approaches could be
explored in future work.

Furthermore, if the model misinterprets constraints in the informal description
of the search problem, such errors can lead to contradictory or otherwise
unsatisfiable constraints. This is detected during the constraint solver
invocation by CBMC, and if detected we again just revert to the initial step of
the process. In future work we might again explore whether the model can recover
from this, if provided this information, or whether it is preferable to just
restart the process, as is currently the case.

Similar to unsatisfiable constraints, the model might also commit formalisation
mistakes that loosen the requirements and thus allow for more solutions than
should actually be the case. The constraint solver can also detect whether two
distinct solutions are possible under the given constraints, and report this
information back to the model. The model could then decide whether the problem
at hand might indeed allow for multiple different solutions, or whether a
correct solution should be unique and thus attempt to correct its mistake.
However, we did not implement this ambiguity detection mechanism in our
prototype. Even though every task in ZebraLogicBench, which we use for our
evaluation, allows for exactly one correct solution, this information is not
shared with the model and is only intended to simplify benchmark result
evaluation.

\section{Experimental Evaluation}
\label{experimental-evaluation}

To evaluate the effectiveness of our approach, we conducted experiments on
ZebraLogicBench, a benchmark suite consisting of 1000 logic grid tasks. In order
to run the evaluation independently, researchers must request access to a
private dataset hosted huggingface.co. We implemented a benchmark runner that
accepts this dataset as input which we share in our open source project
\emph{polymath}. The output result of the benchmark runner is evaluated using
the ZeroEval\footnote{https://github.com/WildEval/ZeroEval} evaluation suite
provided by the benchmark authors.

We use Llama 3.1 70B Instruct for the inference portion in our implementation,
hosted on the Meta-internal MetaGen infrastructure. We run the logic agent
implementation on a developer server with 56 cores and 114GB RAM. In this
environment, running 100 tasks concurrently, full benchmark run takes
approximatley 15 minutes. We evaluted only our own \textbf{Logic Agent}
implementation using this setup. All other results listed were taken directly
from the ZebraLogicBench leaderboard~\cite{zebralogic2024}.

Tab.~\ref{experimental-results-puzzle} lists the performance of our logic agent
implementation compared to other models on the ZebraLogicBench leaderboard. Our
implementation wins the leaderboard with 91.4\% accuracy by a 20\% margin over
OpenAI o1-preview. Notably, o1-preview already scored exceedingly well on the
Easy puzzle category to the point of saturation, and our implementation matches
this high performance. However, an over 28\% gap seperates our implementation
from o1-preview in the Hard puzzle category. This remarkable performance of our
engine is particularly striking when compared to how our base model, LLama 3.1
70B Instruct, performs without the help of a constraint solver. On average, it
reached 24.9\% accuracy across all puzzles, compared to 91.4\% accuracy with the
constraint solver.

\begin{table}
    \resizebox{\linewidth}{!}{
        \begin{tabular}{|l|r|r|r|r|}
            \hline
            \multirow{2}{*}{Model} & \multicolumn{3}{c|}{Puzzle Accuracy} \\
            \cline{2-4}
             & \multicolumn{1}{c|}{All} & \multicolumn{1}{c|}{Easy} &
             \multicolumn{1}{c|}{Hard} \\
            \hline
            \textbf{Logic Agent} & 91.4 & 97.86 & 88.89 \\
            o1-preview-2024-09-12 & 71.4 & 98.57 & 60.83 \\
            o1-mini-2024-09-12-v3 & 59.7 & 86.07 & 49.44 \\
            claude-3-5-sonnet-20241022 & 36.2 & 91.07 & 14.86 \\
            Llama-3.1-405B-Inst-fp8@to\ldots & 32.6 & 87.14 & 11.39 \\
            gpt-4o-2024-08-06 & 31.7 & 84.64 & 11.11 \\
            gemini-1.5-pro-exp-0827 & 30.5 & 79.64 & 11.39 \\
            Llama-3.1-405B-Inst@samba\ldots & 30.1 & 84.64 & 8.89 \\
            Mistral-Large-2 & 29 & 80.36 & 9.03 \\
            claude-3-opus-20240229 & 27 & 78.21 & 7.08 \\
            Qwen2.5-72B-Instruct & 26.6 & 76.43 & 7.22 \\
            gemini-1.5-pro-exp-0801 & 25.2 & 72.5 & 6.81 \\
            Llama-3.1-405B-Inst@hyper\ldots & 25 & 66.67 & 15.38 \\
            gemini-1.5-flash-exp-0827 & 25 & 70.71 & 7.22 \\
            \textbf{Meta-Llama-3.1-70B-Instruct} & 24.9 & 73.57 & 5.97 \\
            \hline
        \end{tabular}
    }
    \caption{ZebraLogicBench Puzzle Accuracy Results}
    \label{experimental-results-puzzle}
\end{table}

A similar result emerges when we evaluate the accuracy of our engine at cell
level. Whereas Tab.~\ref{experimental-results-puzzle} measures how many puzzles
were solved completely in their entirety, Tab.~\ref{experimental-results-cell}
measures how many cells were solved correctly. This measure allows for partial
points, where a model got most of the solution correctly and only committed
smaller mistakes in some sells. Our engine sets a new state-of-the-art in this
category as well, exceeding its base model by a similar margin as in
Tab.\ref{experimental-results-puzzle}.

\begin{table}
    \begin{tabular}{|l||r|}
        \hline
        \multirow{2}{*}{Model} & Cell Ac- \\
        & \multicolumn{1}{l|}{curacy} \\
        \hline
        \textbf{Logic Agent} & 92.98 \\
        o1-preview-2024-09-12 & 75.14 \\
        o1-mini-2024-09-12-v3 & 70.32 \\
        claude-3-5-sonnet-20241022 & 54.27 \\
        Llama-3.1-405B-Inst-fp8@together & 45.8 \\
        gpt-4o-2024-08-06 & 50.34 \\
        gemini-1.5-pro-exp-0827 & 50.84 \\
        Llama-3.1-405B-Inst@sambanova & 39.06 \\
        Mistral-Large-2 & 47.64 \\
        claude-3-opus-20240229 & 48.91 \\
        Qwen2.5-72B-Instruct & 40.92 \\
        gemini-1.5-pro-exp-0801 & 48.5 \\
        Llama-3.1-405B-Inst@hyperbolic & 46.62 \\
        gemini-1.5-flash-exp-0827 & 43.56 \\
        \textbf{Meta-Llama-3.1-70B-Instruct} & 27.98 \\
        \hline
    \end{tabular}
    \caption{ZebraLogicBench Cell Accuracy Results}
    \label{experimental-results-cell}
\end{table}

\section{Threats to Validty}

While we believe we made a convincing case in Sec.~\ref{experimental-evaluation}
that auxiliary tools can significantly boost the performance of state-of-the-art
LLMs on challenging tasks, we would like to address potential threats to the
validity of our results.

Firstly, our engine has thus far only been evaluated against a particular
category of search-based problems in the form of logic grid puzzles. Our
technique may prove less impactful when applied to other search-based problems
formalised by an LLM. This may be due to the fact that our \logiclang\ prototype
is currently incomplete and does not yet support some common Python language and
standard library features, or more fundamentally, that the formalisation of
other search-based problems may not prove as tractable for state-of-the-art LLMs
as logic grid puzzles.

Furthermore, while our approach of using a tailored DSL to allow the LLM to
operate an auxiliary tool yielded excellent results, further experiments will
yet have to show whether this approach can be generalised to other categories of
auxiliary tools and problem domains, such as numerical computational programming
languages or algebraic modeling languages to improve math capabilities of LLMs.
We intend to pursue these problem domains in future work.

\section{Conclusions and Future Work}

In this paper, we presented a novel approach to solving search-based problems
using large language models. By introducing the domain-specific language
\logiclang\ and implementing an agentic solver engine, we demonstrated
significant improvements in performance on the logic puzzle benchmark
\textit{ZebraLogicBench}. Our results show that combining the strengths of LLMs
with those of constraint solvers can lead to remarkable advancements in solving
traditionally challenging tasks.

The success of our approach highlights the potential for further research in
this area. Some promising directions for future work include:

\begin{enumerate}
    \item \textbf{Extending} \logiclang: Developing a more comprehensive and
expressive version of \logiclang\ could enable the formalisation of a wider
range of search-based problems.
    \item \textbf{Improving the Logic Agent}: Enhancing the agentic solver
engine to better handle complex problem statements and iterate on its mistakes
(e.g. syntax errors) could lead to further performance gains.
    \item \textbf{Applying the approach to other domains}: Exploring the
application of our method to other, such as optimisation or mathematical
rasoning, could reveal new opportunities for improvement.
    \item \textbf{Investigating the role of LLMs in problem formalisation}:
Further research into the capabilities and limitations of LLMs in formalising
search-based problems could provide valuable insights into the design of more
effective problem-solving systems.
\item \textbf{Teaching neural nets to search like solvers}: Our translations leverage powerful aspects of constraint solvers: proof checking and search. It would be possible to decompose these strengths. Solvers like Z3 have evolved subtle heuristics for approaching computationally intractable problems that are NP-hard and more, and if we can train nets in a way that learns these heuristics they might provide benefit more broadly than as the targets of translation-based work. In this context, work such as this could provide baselines or targets for what we would want out of the training.

\end{enumerate}

In particular, we are currently working on a DSL suitable for tackling first
order logic problems and intend to evalute this engine against the FOLIO
benchmark. \cite{han2022folio}

Building on this project, our broader research aims to further explore the
potential of combining large language models with specialized reasoning tools.
By pursuing these avenues of research, we believe that it is possible to develop
even more powerful and efficient problem-solving systems.

\appendix

\bibliographystyle{named}
\bibliography{references}

\end{document}